\documentclass[10pt,twocolumn,letterpaper]{article}

\usepackage{iccv}
\usepackage{times}
\usepackage{epsfig}
\usepackage{graphicx}
\usepackage{amsmath}
\usepackage{amssymb}

\usepackage{multirow}

\newcommand{\Tokyo}{Tokyo~24/7\xspace}

\newcommand{\comment}[1]{}
\renewcommand{\paragraph}[1]{\vspace{.5\baselineskip}\noindent{\bf #1}\xspace}

\def\roxf{$\mathcal{R}$Oxf\xspace}
\def\rpar{$\mathcal{R}$Par\xspace}

\newcommand{\mdl}[1]{``{#1}''\xspace}

\usepackage[breaklinks=true,bookmarks=false]{hyperref}

\iccvfinalcopy

\def\iccvPaperID{4874}

\ificcvfinal\fi

\begin{document}

\title{No Fear of the Dark: \\
{\large Image Retrieval under Varying Illumination Conditions}}

\author{
Tomas Jenicek \qquad Ond{\v r}ej Chum\\
Visual Recognition Group, Faculty of Electrical Engineering, Czech Technical University in Prague\\
}

\maketitle
\ificcvfinal\thispagestyle{empty}\fi

\begin{abstract}
Image retrieval under varying illumination conditions, such as day and night images, is addressed by image preprocessing, both hand-crafted and learned. Prior to extracting image descriptors by a convolutional neural network, images are photometrically normalised in order to reduce the descriptor sensitivity to illumination changes.
We propose a learnable normalisation based on the U-Net architecture, which is trained on a combination of single-camera multi-exposure images and a newly constructed collection of similar views of landmarks during day and night.
We experimentally show that both hand-crafted normalisation based on local histogram equalisation and the learnable normalisation outperform standard approaches in varying illumination conditions, while staying on par with the state-of-the-art methods on daylight illumination benchmarks, such as Oxford or Paris datasets.

\end{abstract}

\section{Introduction}

Since the first successful image retrieval methods~\cite{Sivic-ICCV03,Nister-CVPR06}, the field went through a rapid development. Numerous methods based on local features~\cite{Mikolajczyk-IJCV04,Matas-BMVC02} and their descriptors~\cite{Lowe-IJCV04} were improved in many directions, including spatial verification~\cite{Philbin-CVPR07,Jegou-ECCV08,Perdoch-CVPR09}, descriptor aggregation~\cite{Jegou-CVPR10,Perronnin-CVPR10}, and convolutional neural network (CNN) based feature detectors~\cite{Yi-ECCV16} and descriptors~\cite{Tian-CVPR17,Mishchuk-NIPS17}. Recently, image retrieval approaches based on global CNN descriptors~\cite{Arandjelovic-CVPR16,Gordo-ECCV16,Radenovic-TPAMI18} started to dominate due to their efficiency both in the search time and memory footprint. 

The challenges of image and particular object retrieval lie mainly in increasing the efficiency for large collections of images and in improving the quality of retrieved results. Scaling up to very large collections of images is addressed by efficient extraction of global CNN features and consequent efficient encoding~\cite{Jegou-PAMI11} and nearest neighbour search~\cite{Babenko-CVPR16,Johnson-17}.
Another direction of research considers retrieval of instances that exhibit significant geometric and/or photometric changes with respect to the query.

Various types of geometric changes appear in image collections, for example change of scale, such as when the query object covers only a small part of the database image, change in the view-point, and severe occlusion. Methods based on local features and efficient geometric verification~\cite{Stewenius-ECCV12} have shown good retrieval performance on significant geometric changes~\cite{Mikulik-SISAP13,Mikulik-ACCV14}.

\begin{figure}[t] \centering
    \includegraphics[scale=0.26]{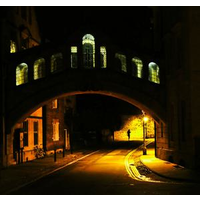}
    \hspace{0.6cm}
    \includegraphics[scale=0.26]{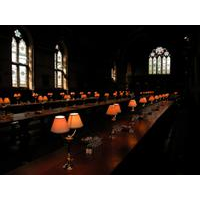}
    \includegraphics[scale=0.26]{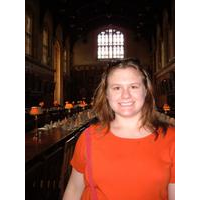}
    \includegraphics[scale=0.26]{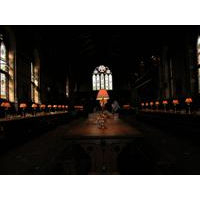}
\\
    \includegraphics[scale=0.26]{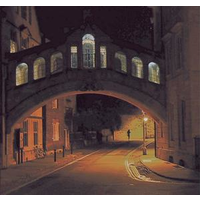} \hspace{0.6cm}
    \includegraphics[scale=0.26]{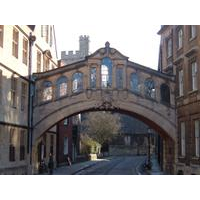}
    \includegraphics[scale=0.26]{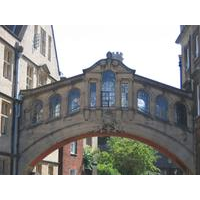}
    \includegraphics[scale=0.26]{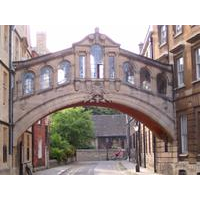}\\[2pt]

    \caption{An example of a night query where learned photometric normalisation improves the results of image retrieval. For a query image (top left), images from Oxford 5k~\cite{Radenovic-CVPR18} retrieved by VGG GeM~\cite{Radenovic-TPAMI18} are shown (top row). When using a learned normalisation, the query image is first normalised (bottom left) and then used to retrieve images using the same procedure (bottom row).}
\label{fig:motivation}
\end{figure}

Image retrieval with photometric changes is partially addressed by local-feature based approaches, as the local-feature descriptor extraction typically contains a local photometric normalisation step. It has been shown, \eg in~\cite{Radenovic-CVPR16}, that local features are able to connect day and night images through (a sequence of) images with gradual change of illumination. 
For CNN based approaches, it has been shown that the state-of-the-art methods fail under severe illumination changes, even though relevant information is preserved (\eg in the form of edges)~\cite{Radenovic-ECCV18}.
This can be attributed to the lack of training data, as it is difficult to obtain large amount of day-night image pairs in sufficient quality and diversity. In this paper, we address CNN-based image retrieval with significant photometric changes. The goal is to provide a mapping from images to a descriptor space, where nearest neighbour search will be capable of retrieving instances with significantly different illumination. At the same time, the performance on day-to-day retrieval should remain competitive with the state of the art. In other words, we are interested in a method that works under all illumination conditions, see Fig.~\ref{fig:motivation}.

\begin{figure}[t] \centering
    \includegraphics[width=\columnwidth]{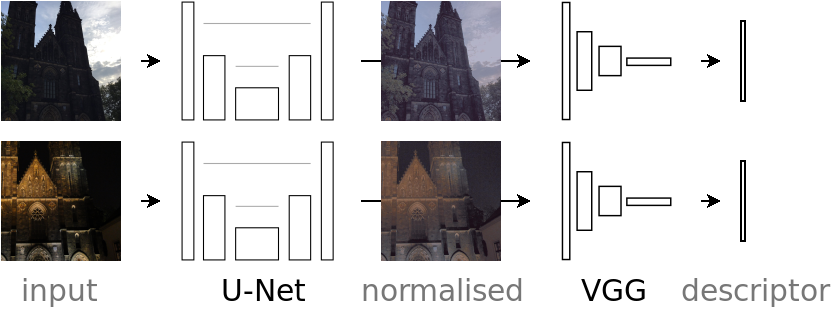}\\[3pt]

    \caption{For fine-tuning, the normalisation network (U-Net) is prepended to the embedding network (VGG) and both are trained in a Siamese manner on pairs of images. Each image of the input pair is first normalised and then embedded. A contrastive loss is applied to the distance between resulting descriptors.}
\label{fig:finetune}
\end{figure}

In this work, we propose to perform a photometric normalisation that preprocesses the images (both the query and the database images) before extracting the descriptors.
The goal of this stage is to enhance the discriminative information in images taken under challenging illumination conditions and to bring them closer to typical daylight images. 
We investigate various types of hand-crafted normalisation operating both globally and locally on the image. 
We also design a neural network and train it to transform an image to match given statistics. The network is pre-trained on a collection of multi-exposure photographs~\cite{Chen-CVPR18}. Multi-exposure images are relatively easy to collect, as opposed to aligned day and night images without significant changes in the scene. For fine-tuning, the photometric normalisation is pre-pended to the embedding network and trained in an end-to-end manner with contrastive loss, see Fig.~\ref{fig:finetune}.
The proposed normalisation methods are compared with a number of different approaches including edge map extraction, which is considered partially illumination invariant~\cite{Radenovic-ECCV18}.

The main contribution of this paper is the introduction of the normalisation step. We propose performing photometric normalisation prior to extracting the descriptors. Both hand-crafted and learned normalisation is evaluated. We construct a training day-night dataset from existing 3D reconstructions which was made publicly available. Both the proposed normalisation and the constructed dataset is experimentally shown to improve the performance on challenging queries.

\section{Related work}

To reduce the sensitivity of local feature descriptors to illumination changes, an intensity normalisation step is introduced to the descriptor generation process, as in one of the most popular descriptors, SIFT~\cite{Lowe-IJCV04}. Another approach is based on geometric hashing~\cite{Lamdan-ICCV88,Chum-CVPR06} where the feature descriptor is not based on the appearance but on mutual positions of near-by features. 

Approaches making the local-feature descriptor insensitive to illumination changes alone are not sufficient to match difficult image pairs, as they rely on the feature detector to fire at the same locations despite the illumination change. One of the first approaches to learn illumination invariant feature detector was a Temporally Invariant Learned DEtector (TILDE)~\cite{Verdie-CVPR15}. In TILDE, the detector is trained on a dataset of images from 6 different scenes collected over time by still web cameras pointing out of a window. First, feature point candidates are selected. The selection criterion is stability across a number of roughly aligned webcam images collected over time. A regressor giving high responses in the candidate locations and low everywhere else is learned. 

The problem of day and night visual self-localisation using GPS-annotated Google StreetView images is addressed in~\cite{Torii-CVPR2015}.
The \Tokyo dataset of day, sunset and night images taken by a cell phone camera is used for query images. The authors demonstrate that for a dense VLAD descriptor~\cite{Jegou-CVPR10}, matching across large changes in the scene appearance becomes much easier when both the query image and the database image depict the scene from approximately the same viewpoint. To perform the visual localisation, StreetView panoramas and corresponding depth-maps are used to render a large number of virtual views by ray-tracing with view-points on a 5m $\times$ 5m grid and 12 view directions at every view-point. Significant boost in performance is achieved when the queries are matched against the virtual views rather than the original panoramas.
The \Tokyo dataset is described in more detail in section~\ref{sec:experiments} as we use it for evaluation.

EdgeMAC~\cite{Radenovic-ECCV18} performs reasonable image matching in the presence of a significant change in illumination, especially when the colours and textures are corrupted. However, for a standard imaginary, dropping all the information but edges certainly degrades the performance, as already observed by \cite{Radenovic-ECCV18} and confirmed by our experiments.

Methods enhancing visual quality of images taken under bad light conditions were proposed. In~\cite{Chen-CVPR18}, raw output from the image sensor is taken and a neural network is used to enhance the visual appearance, as if the image was taken with long exposure. Camera (sensor) dependent models are learned from a dataset of multiple-exposure images of static scenes with qualitatively very impressive results.

\section{Photometric normalisation}

Image descriptors for image retrieval are extracted by a system of two components: photometric normalisation and embedding network. The normalisation translates images to an image domain less sensitive to illumination changes. The embedding network provides the mapping from the image to the descriptor space, in which  nearest neighbour search is used to retrieve similar images.
Two types of photometric normalisation are investigated, image preprocessing by hand-crafted normalisation and a normalisation network prepended to the embedding network.
 
\subsection{Hand-crafted normalisation}

Hand-crafted normalisation, specifically {\bf histogram equalisation}, {\bf CLAHE} and {\bf gamma correction}, is tested first in order to evaluate the need for a learnable normalisation network. We refer to~\cite{szeliski2010computer} for detailed  description of the algorithms. In the proposed pipeline, the image to be normalised is converted from RGB to LAB colour space, intensity transformation is applied on the lightness channel, and the image is converted back to RGB colour space before being used as an input to the embedding network.

In histogram equalisation, monotonic pixel intensity mapping is found, so that the histogram of mapped intensities is flat. 

In adaptive histogram equalisation, the image is divided into non-overlapping blocks and histogram equalisation is performed on each block independently. Each pixel intensity is then bilinearly interpolated from the four closest block mapped intensities, making transitions between blocks smooth. 
When a contrast limit is applied, the original histogram is mapped to a clipped histogram, which is not uniformly distributed in general.
The clipped histogram is constructed from the original histogram by uniformly redistributing pixels from the frequent intensity bins (bins whose value exceeds the clip limit)~\cite{szeliski2010computer}.
With clip limit equal to 1, the resulting histogram is flat, so the result is identical to histogram equalisation. The Contrast Limited Adaptive Histogram Equalisation (CLAHE) is a combination of all the techniques described above.

In gamma correction, pixel values in the range between 0 and 1 are raised to the power of chosen positive exponent. The exponent in gamma correction is chosen for each image such that the corrected image mean is equal to the dataset average. This is performed via a fast secant method that allows to perform it during image loading.

\paragraph{Implementation details.} For CLAHE, each image is split into a grid of 8x8 windows, so that the longer side of each window is 45px. The clip limit is set to 4 for all experiments which consistently yielded the best results.

\subsection{Learnable normalisation}

In this section, the architecture of the normalisation network is described and the details of its separate pre-training, including the description of the dataset used, are given.

\begin{figure*} \centering
    \includegraphics[width=4.1cm]{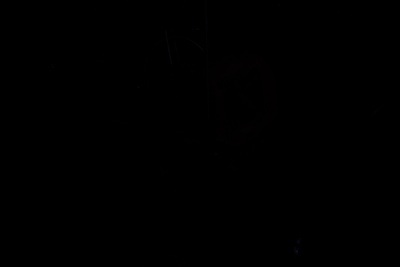}
    \includegraphics[width=4.1cm]{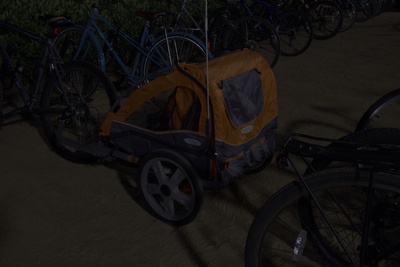}
    \includegraphics[width=4.1cm]{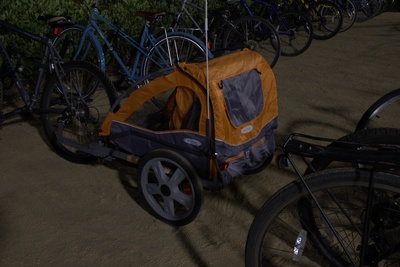}
    \includegraphics[width=4.1cm]{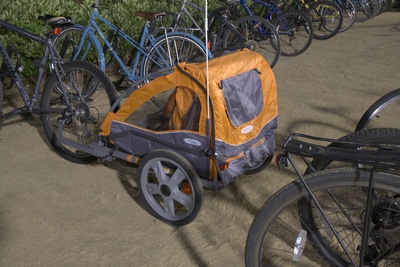}\\[1pt]

    \caption{Example images from dataset See in the dark~\cite{Chen-CVPR18} used in training. From left to right: short exposure, interpolated, long exposure and extrapolated image. The first and third image is from the dataset, the second and fourth is synthesised.}
\label{fig:dataset_sid}
\end{figure*}

\begin{figure*} \centering
    \parbox[b]{.49\linewidth}{\centering%
    \includegraphics[width=3.8cm]{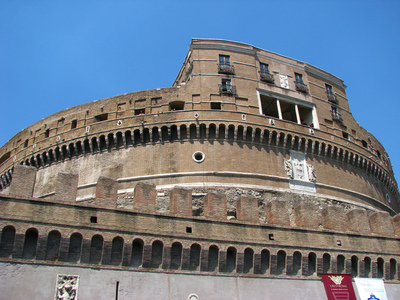}
    \includegraphics[width=3.8cm]{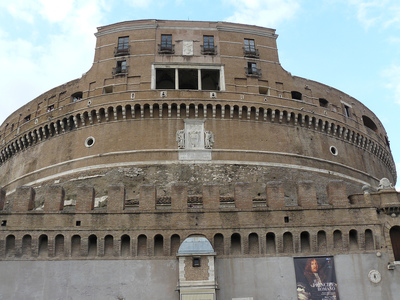}\\
    \includegraphics[width=3.8cm]{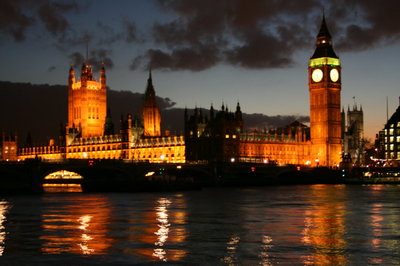}
    \includegraphics[width=3.8cm]{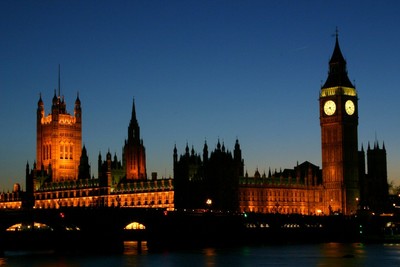}\\
    (a)}\hfill
    \parbox[b]{.49\linewidth}{\centering%
    \includegraphics[width=3.8cm]{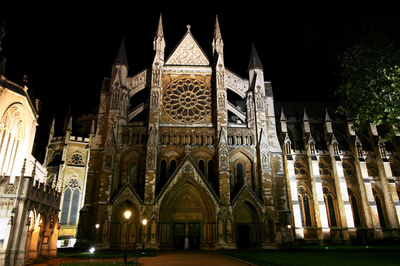}
    \includegraphics[width=3.8cm]{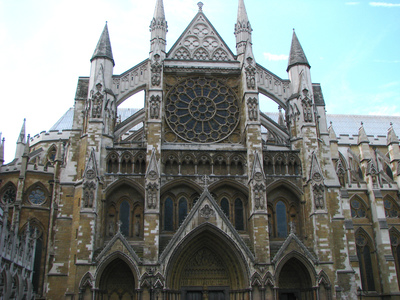}\\
    \includegraphics[width=3.8cm]{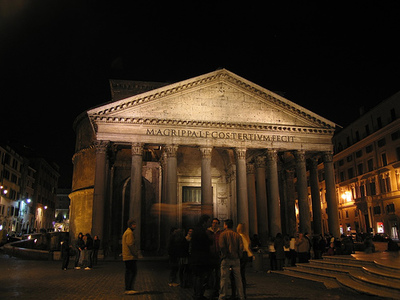}
    \includegraphics[width=3.8cm]{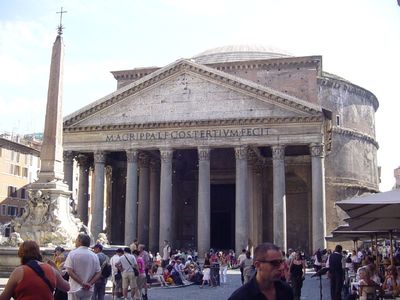}\\
    (b)}

    \caption{Examples of positive image pairs obtained from a 3D structure-from-motion model. The left image is an anchor, the right a hard-positive example used for training with two different datasets: (a) Retrieval-SfM~\cite{Radenovic-TPAMI18}, (b) Retrieval-SfM-N/D.}
\label{fig:dataset_sfm}
\end{figure*}

\subsubsection{Architecture}

The {\bf normalisation network} is designed to transform an image into a pixel-aligned image with different image statistics. The input to the normalisation network consists of the RGB channels of the input image and a lightness channel matching the target image statistics. This additional channel is obtained by transforming the input lightness channel to the target lightness channel histogram by histogram matching, all in the LAB colour space. The output of the network is an RGB image.

The normalisation network has the U-Net architecture~\cite{ronneberger2015u}, in particular, the implementation is adopted from~\cite{isola2017image}. The network architecture from~\cite{isola2017image} was altered for the normalisation task. After the last transposed convolution, the tanh layer is replaced by a ReLU layer followed by a convolution with 32 input channels and 3 output channels. The number of output channels of the last transposed convolution was changed accordingly. In order to improve the performance, batch-norm layers were removed and bias was added to all convolutions.
Each individual adaptation has increased the performance on the task of mapping across different exposure times, measured on the validation set. The original U-Net architecture~\cite{ronneberger2015u} performed similarly to the adapted architecture of~\cite{isola2017image} but with higher GPU memory and time requirements. In our experiments, specifically the U-Net jointly scenario, the increase was from 4.1GB, 5hrs to 11.6GB, 11hrs with a performance gain of less than 1\% on average.

The use of a lightness channel from the LAB colour space is a design choice that provided slightly better results than corresponding channel from LUV, HLS, HSV and RGB average in preliminary experiments. It is also possible to add the unaltered input image lightness channel to the input of normalisation network. It marginally increases the performance, but the improvement is not consistent and is less than 1\% on average, so the simpler network architecture is reported.

\paragraph{Implementation details.} Due to the U-Net architecture, used for the normalisation network, both input image dimensions must be divisible by 256. During pre-training, images are down-scaled and/or cropped to meet this requirement. During fine-tuning and for inference, images are padded to the smallest larger dimensions divisible by 256, if necessary. To maximise contextual information at the border, reflection padding is used. After normalising the image, the padding is removed, so that the output image of the normalisation network has the same dimensions as the input image.

During pre-training of the normalisation network, the target statistics are extracted from target ground truth images through histogram matching. When the normalisation network is prepended to the embedding network, histogram equalisation is performed instead. The equalised histogram matches very closely the average image lightness distribution which we empirically verified on the Retrieval-SfM dataset~\cite{Radenovic-TPAMI18}.

\subsubsection{Pre-training dataset}
{\bf See-in-the-dark} dataset (SID) is used to pre-train the normalisation network. It was introduced by~\cite{Chen-CVPR18} for the task of enhancing (raw) images captured with extremely low exposure time. This dataset consists of 424 different static scenes, both indoors and outdoors, taken by two different cameras with different sensors - Sony~$\alpha$75~II and Fujifilm~X-T2 with the resolution of $4240 \times 2832$ and $6000 \times 4000$ pixels respectively. Each scene was captured repeatedly in low light conditions in a number of short-exposure times and one long-exposure time. For each scene, a pair of long- and short-exposure images is selected. If multiple short-exposure images are available, the one with the longest exposure time is picked. This yields 827 precisely pixel-to-pixel aligned low and high-exposure image pairs.

Two types of data augmentation are used with on dataset. First, the high resolution of the images allows for re-scaling and cropping. The images are split into $2127 \times 1423$ and $2010 \times 1343$ patches for Sony and Fujifilm camera respectively. This enables combining patches from multiple images and cameras in a single batch without the overhead of reading images in their original size. The patches are scaled by a random factor between 0.4 and 0.8 to reduce the noise, then randomly horizontally flipped and randomly cropped to the final size of $768 \times 512$. For validation, only a single centre crop is performed. As another data augmentation, additional illumination levels are synthesised from the raw images. For each aligned pair of images, raw sensor data are processed using the standard pipeline~\cite{szeliski2010computer} and before applying the gamma function, pixel values for two different exposure times are interpolated by a linear function. This models the amount of light hitting the sensor and allows to extrapolate images with illumination levels not present in the original dataset. There are 3 interpolated and 2 extrapolated illumination levels synthesised; the short exposure image is never used, as there is no signal in the RGB image. Example images are shown in Fig.~\ref{fig:dataset_sid}.

\begin{figure}[t] \centering
    \includegraphics[width=\columnwidth]{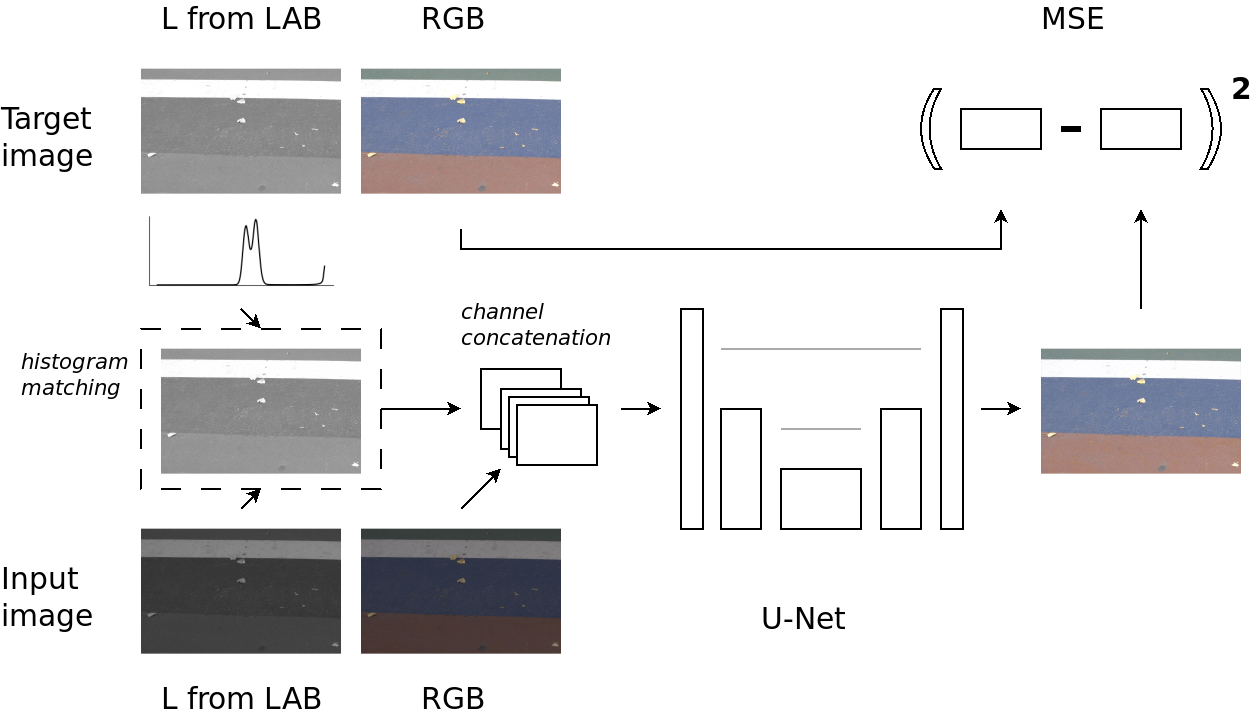}\\[4pt]

    \caption{Pre-training of the normalisation network on pixel-aligned image pairs. Each image pair is converted from RGB to LAB from which only the lightness channel is kept. The input image lightness channel (bottom-left) is transformed to the statistics of the target image lightness channel (top-left) through histogram matching. The resulting channel is concatenated with the input image RGB channels and fed to the normalisation network (U-Net). The loss function (MSE) is computed between the RGB image outputted by the network and the target RGB image.}
\label{fig:pretrain}
\end{figure}

\subsubsection{Pre-training}

The normalisation network is first trained on pixel-aligned pairs of images taken in different illumination conditions. The goal is, given one of the images (input) and statistics of the other (target) to reconstruct the target image. For the embedding network, we use the off-the-shelf pre-trained VGG retrieval CNN provided by~\cite{Radenovic-TPAMI18}.

Pre-training of the normalisation network is performed using the See-in-the-dark dataset. A pair of input and target image is chosen randomly from a set of images of each scene. No constraints are set on the pair, the target image can have both longer or shorter exposure time than the input image. The pre-training is summarised in Fig.~\ref{fig:pretrain}.

The loss function is the mean squared error between the predicted and target image, computed over all pixels. The network is trained for 44 thousand iterations with a batch size equal to 5. An SGD optimiser with learning rate of 0.001, momentum 0.1 and weight decay $10^{-4}$ is used.

\section{Fine-tuning for retrieval}
The proposed illumination-invariant retrieval method is fine-tuned in a two-stage process.
In the first stage, the embedding network is fine-tuned separately, without normalisation. In all experiments, the VGG network architecture with GeM pooling as provided by the authors of~\cite{Radenovic-TPAMI18} is used. The network is trained from off-the-shelf classification network~\cite{SZ14} minimising the contrastive loss on the image descriptors, following the procedure of~\cite{Radenovic-TPAMI18}.
In the second stage, normalisation is prepended to the embedding network and the final composition is fine-tuned also using the contrastive loss and in the same setup. This is common for both hand-crafted and learnable normalisation. In case of learnable normalisation, different scenarios are distinguished based on which network is trained.
A common practice in image retrieval is to apply whitening on the image descriptors extracted by the embedding network. Specific whitening is learned for each trained network following the procedure of~\cite{Radenovic-TPAMI18}. In all our experiments, retrieval is performed with whitened descriptors.

\subsection{Training datasets}
Two datasets were used to fine-tune our network - one of them is publicly available, the other one is newly created. In the following, we provide an overview of these datasets. Example images of these datasets are shown in Fig.~\ref{fig:dataset_sfm}.

\paragraph{Retrieval-SfM} dataset is used in~\cite{Radenovic-TPAMI18} to fine-tune a CNN for retrieval. We use the predefined geometrically validated image clusters and hard negative mining procedure as described in~\cite{Radenovic-TPAMI18}. However, most of the selected anchor and positive images are pairs of daylight images, occasionally a pair of night images is included, see Fig.~\ref{fig:dataset_sfm}~(a). 

\paragraph{Retrieval-SfM-N/D} is a novel dataset constructed from the same 3D reconstruction as Retrieval-SfM. We extracted hard positive image pairs with different lighting conditions, these hard positives are complementary to those provided in Retrieval-SfM. Example images are shown in Fig.~\ref{fig:dataset_sfm} (b). This dataset was made available on the project web page\footnote{http://cmp.felk.cvut.cz/daynightretrieval}.

In~\cite{Radenovic-TPAMI18}, in order to ensure the same surface is visible in positive image pairs, a certain number of features reconstructed to a common 3D point is required. However, even two geometrically very similar views with significant change in illumination may share only a small number of matching SIFT features. To find images observing the same scene surface, we approximate the surface visible in an image by a ball. The centre of this ball is equal to the mean of 3D points reconstructed for an image and the radius is given by a standard deviation of those points. To validate that two images depict the same part of the surface, the intersection over union of corresponding balls must be greater than 0.55. Furthermore, for a positive image pair, the angle between estimated camera optical axes is limited to 45 degrees.
The (relatively rough) ball approximation followed by volume intersection over union measure is very fast and exhaustively applicable to even large 3D models, providing satisfactory results for a wide range of objects without obvious false positives.

The procedure above assigns to each image participating in a 3D model a list of potential positive images.
The hard positive image pairs are chosen so that they maximise the difference in illumination among geometrically similar images. We measure the illumination difference as the difference in a trimmed-mean value of lightness in the LAB colour space where the lightest 40\% and darkest 40\% of pixels are dropped. This measure is robust to the presence of image frames, large occluding objects, \etc, which can be either light or dark.

We have constructed 20 thousand illumination-hard-positive image pairs with the largest difference in the illumination. The anchor image is chosen to be the darker image and a positive example the lighter. For the anchor images, a standard hard negative mining is performed during training~\cite{Radenovic-TPAMI18}. 

\subsection{Fine-tuning}

To fine-tune the composition of normalisation and embedding network, three approaches are compared. First, the embedding network is frozen and only the normalisation network is fine-tuned for retrieval. Second, the normalisation network is frozen, and the embedding network is trained. Finally, both networks are trained jointly with alternating update of the normalisation and the embedding network. All three approaches are trained on a mixture of Retrieval-SfM and Retrieval-SfM-N/D hard positives and mined hard negatives and this mixture is also used for consequent whitening.

In all three approaches, the training procedure of~\cite{Radenovic-TPAMI18} is followed. The training is performed for 4 thousand iterations, 10 epochs of 400 iterations each, with a batch size of 5. All images are downscaled to have the longer edge equal to 362 px for training and to 1024 px for validation. For each anchor image, five hard negative images are mined at the beginning of each epoch. In each epoch, hard negatives for 2 thousand query images are mined from the pool of 20 thousand images. The margin in the contrastive loss is set to 0.75.

\paragraph{Fine-tuning} of the {\bf normalisation} network. The gradient from the contrastive loss is backpropagated through the embedding network to the normalisation network. Weights of the embedding network are not updated during backpropagation, treating the embedding network as a loss function of the normalisation network. The learning parameters for the normalisation network remain the same as in pre-training.

\begin{figure}[t]
{\centering
    \includegraphics[width=.326\linewidth]{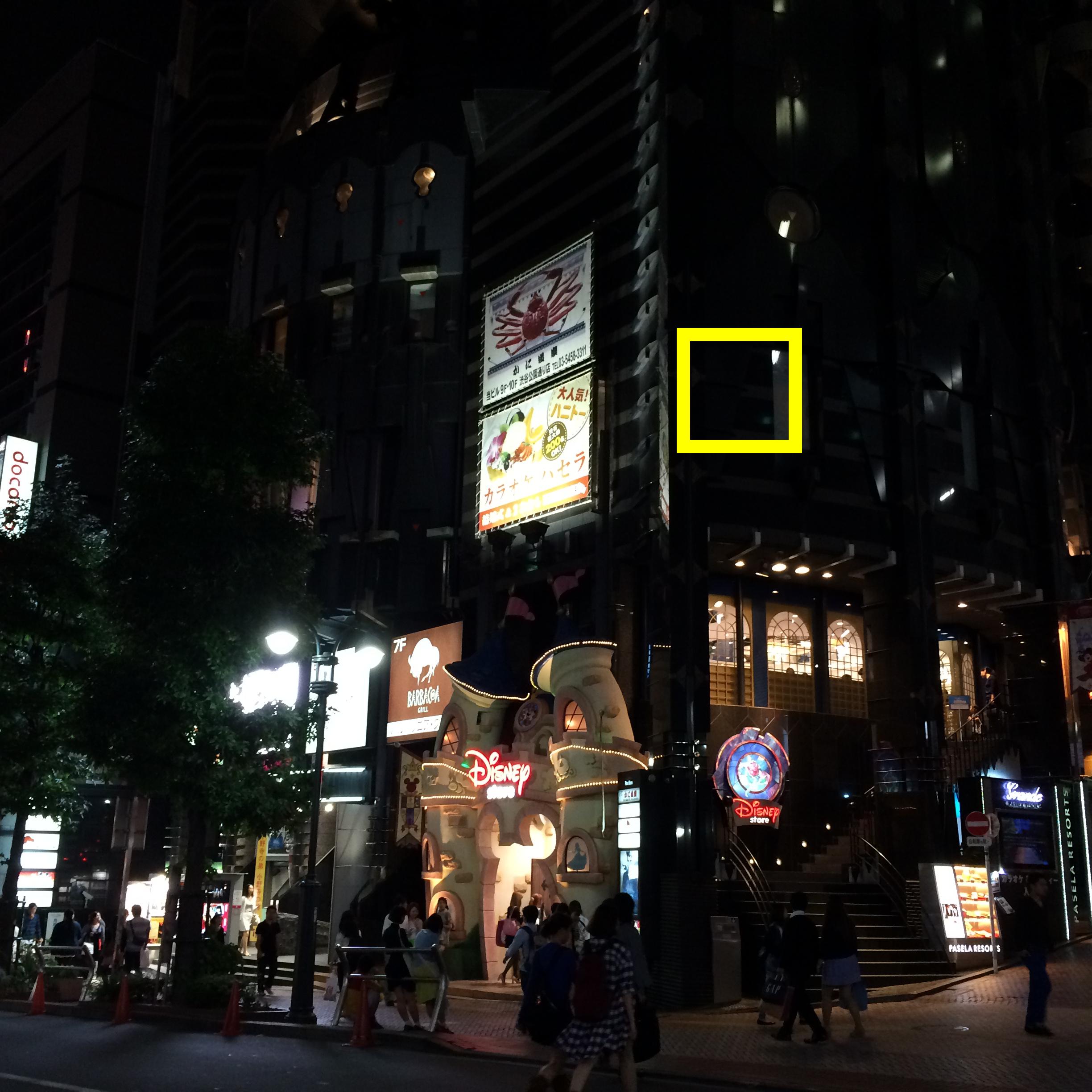}
    \includegraphics[width=.326\linewidth]{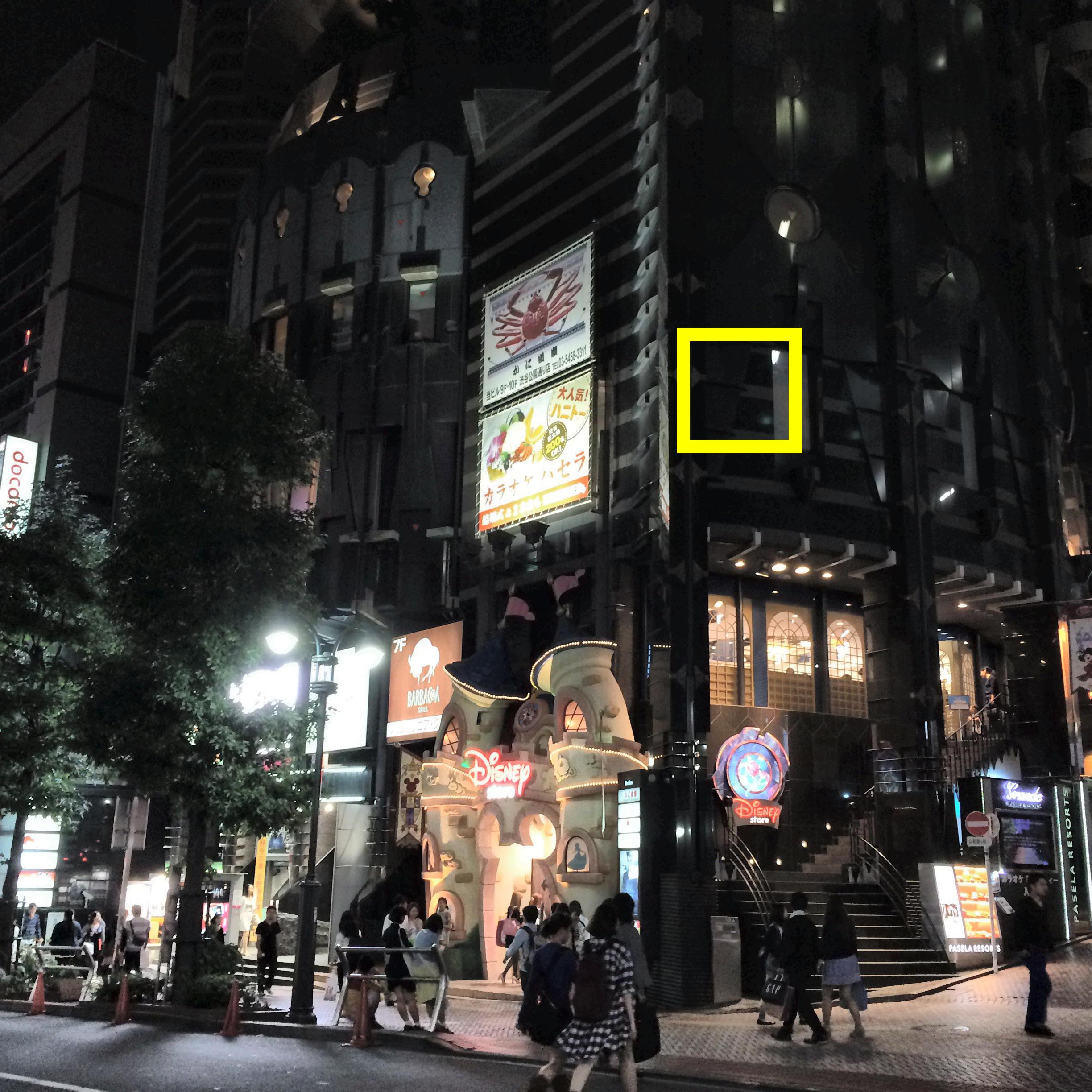}
    \includegraphics[width=.326\linewidth]{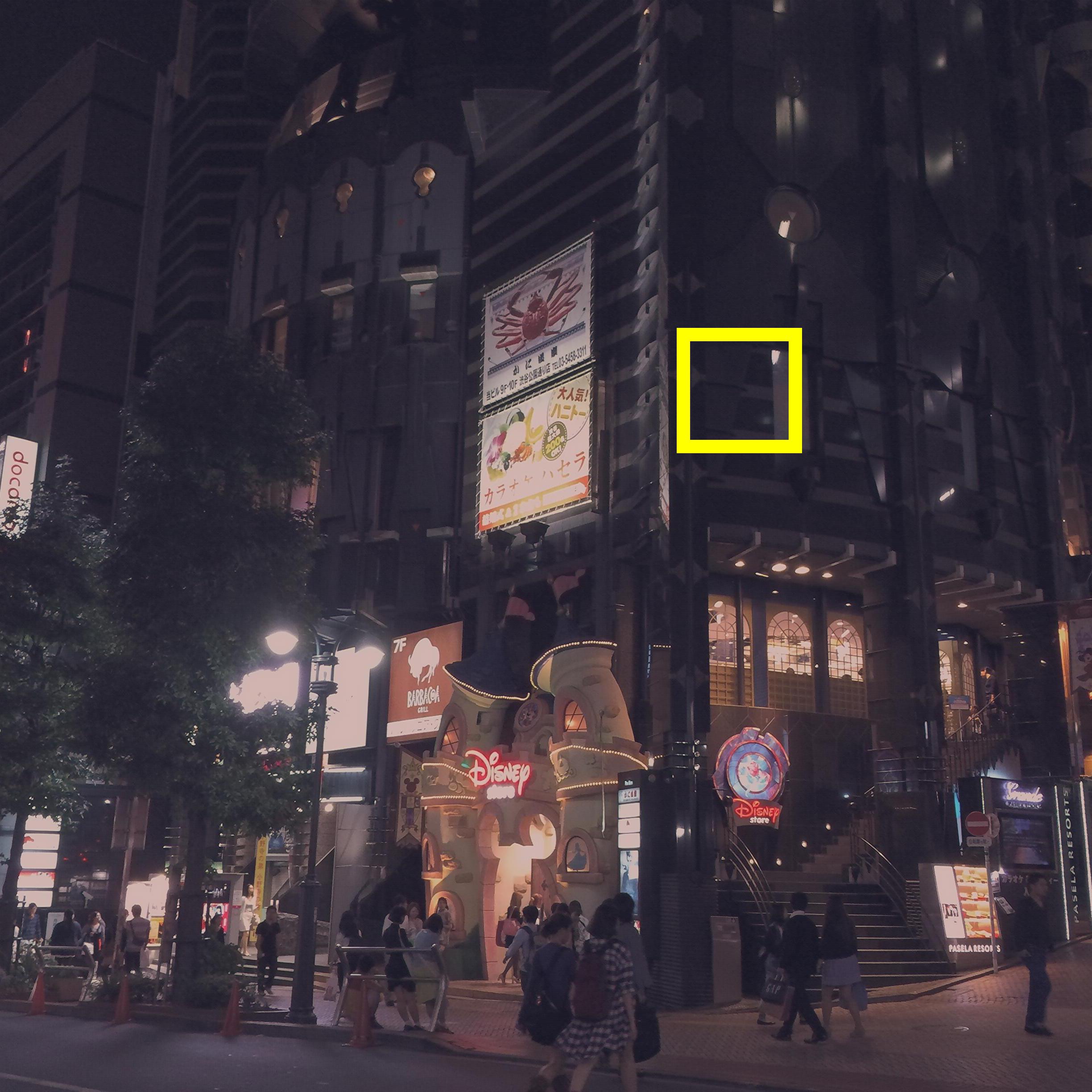}
    
    \includegraphics[width=.326\linewidth]{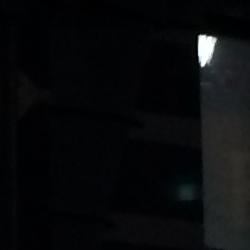}
    \includegraphics[width=.326\linewidth]{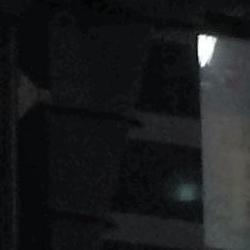}
    \includegraphics[width=.326\linewidth]{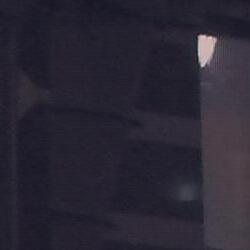}

    \caption{Unaltered image from Tokyo 24/7 dataset (left), normalised by CLAHE (middle) and by U-Net from U-Net jointly N/D model (right). Best viewed on a computer screen.}
\label{fig:qualitative}}
\end{figure}

\paragraph{Fine-tuning} of the {\bf embedding} network is performed with the Adam optimiser with a learning rate of $10^{-6}$, weight decay of $10^{-4}$ and momentum parameters $\beta_1 = 0.9$ and $\beta_2 = 0.999$~\cite{Radenovic-TPAMI18}.

\paragraph{Joint fine-tuning} uses a separate optimiser for each network, due to the sensitivity of both U-Net and pre-trained VGG to the optimiser choice. SGD is used to update the normalisation network while Adam is used to update the embedding network. The parameters for each optimiser are the same as in normalisation and embedding network fine-tuning. The training updates weights of only one network at a time, alternating the networks every 10 iterations.

\section{Experiments} \label{sec:experiments}

To evaluate the effect of the proposed image normalisation, we test all methods on two standard benchmarks for image retrieval, and propose a new evaluation protocol for image retrieval with severe illumination changes. We compare hand-crafted and learned normalisation with state-of-the-art baselines. The effects of hand-crafted and learned normalisation are visualised in Fig.~\ref{fig:qualitative}.

\newcommand{\doubledim}{{\color{red}\bf !}\xspace}
\newcommand{\firstsup}[1]{{\bf\color{red} #1}}
\newcommand{\secondsup}[1]{{\bf #1}}
\setlength{\tabcolsep}{6.8pt}
\begin{table}[t] \centering
\begin{tabular}{|l||r||r|r|r|}\hline
    Method & Avg & Tokyo & \roxf & \rpar \\\hline\hline
    VGG GeM~\cite{Radenovic-TPAMI18} & 69.9 & 79.4 & \firstsup{60.9} & 69.3 \\\hline
    EdgeMAC~\cite{Radenovic-ECCV18} & 45.6 & 75.9 & 17.3 & 43.5 \\\hline\hline
    VGG GeM N/D & 71.1 & 83.5 & 60.0 & 69.8 \\\hline
    EdgeMAC+VGG \doubledim & 71.2 & 85.4 & 59.4 & 68.8 \\\hline\hline
    Gamma corr. N/D & 70.9 & 84.6 & 59.5 & 68.7 \\\hline
    Histogram eq. N/D & 71.6 & \secondsup{86.8} & 59.6 & 68.3 \\\hline
    CLAHE & 71.6 & 84.1 & \secondsup{60.8} & 69.8 \\\hline
    CLAHE N/D & \firstsup{72.4} & \firstsup{87.0} & 60.2 & \firstsup{70.0} \\\hline\hline
    U-Net embed N/D & 70.9 & 86.4 & 58.1 & 68.3\\\hline
    U-Net norm N/D & 71.0 & 83.2 & 60.0 & \secondsup{69.9} \\\hline
    U-Net jointly & 69.8 & 79.8 & 59.9 & 69.7 \\\hline
    U-Net jointly N/D & \secondsup{72.1} & 86.5 & 60.2 & 69.6 \\\hline
\end{tabular}\\[3pt]

    \caption{Comparison of baseline, improved baseline, hand-crafted and learnable normalisation methods (corresponding to visually distinguished blocks) in terms of mAP on \Tokyo, \roxf Medium and \rpar Medium datasets. The average mAP on the three datasets is also reported. Fine-tuning was performed either on the Retrieval-SfM-N/D or Retrieval-SfM dataset. For models based on the learnable normalisation (U-Net), results for three fine-tuning setups (embedding, normalisation, joinlty) are provided where each differ in the network that was fine-tuned. Baselines marked with \doubledim use descriptors of double dimension (i.e. 1024D) compared to others. Best score is emphasised by red bold, second best by bold.}
\label{tab:results}
\end{table}

\subsection{Datasets and evaluation protocol}

\begin{figure}[t] \centering
\newcommand{\includegraphicsx}[1]{\includegraphics[width=2.7cm]{{illustrations/tokio/#1}.jpg}}
    \includegraphicsx{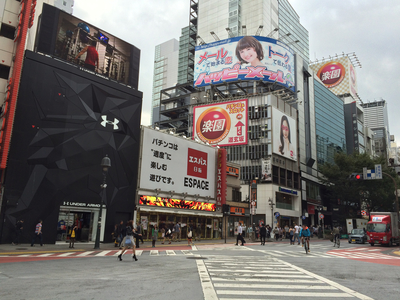}
    \includegraphicsx{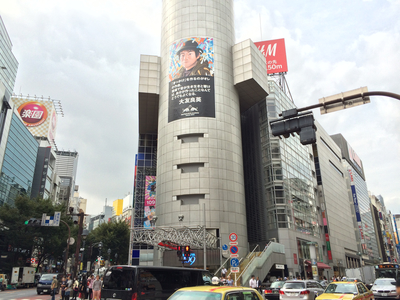}
    \includegraphicsx{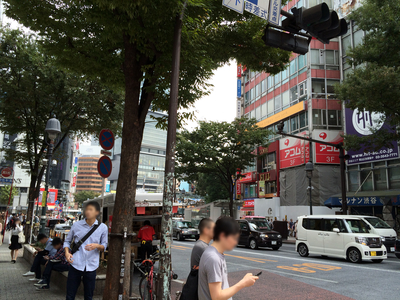}\\
    \includegraphicsx{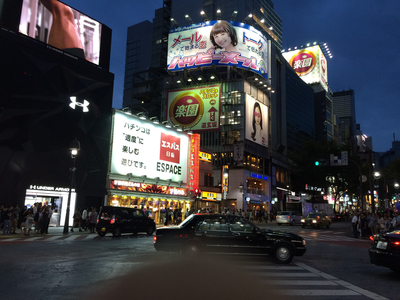}
    \includegraphicsx{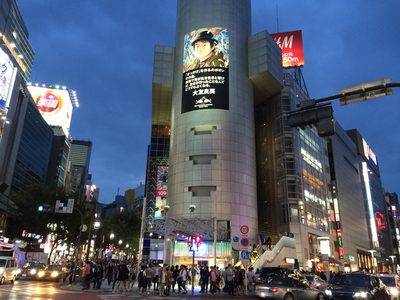}
    \includegraphicsx{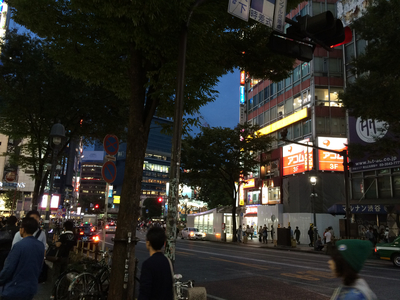}\\
    \includegraphicsx{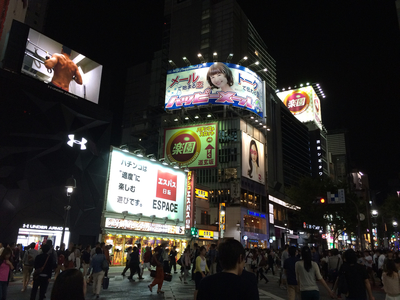}
    \includegraphicsx{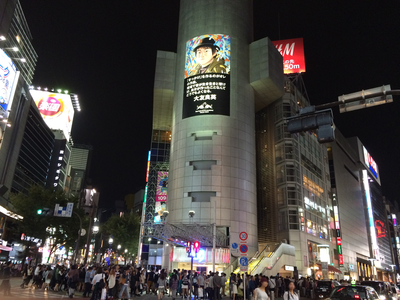}
    \includegraphicsx{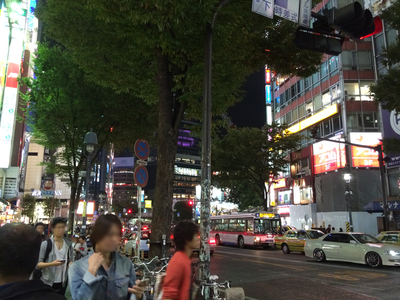}\\

    \caption{A location example from the \Tokyo dataset. Rows represent day, sunset and night light conditions respectively. Columns correspond to different viewing directions. Note the overlap between the first two viewing directions and no overlap between the second and the third.}
\label{fig:tokyo}
\end{figure}

The \Tokyo dataset consists of phone-camera photographs from~\cite{Torii-CVPR2015} taken at 125 locations; the Street View images, used as database images in~\cite{Torii-CVPR2015}, are not included. At each location, images at three different viewing directions were taken at three different light conditions (day, sunset and night). This amounts for nine images per location and 1125 images in total in the dataset. Images taken at different light conditions in the same direction have significant overlap of the photographed surface. However, images from the same location taken in different viewing directions may or may not overlap, as can be seen in Fig.~\ref{fig:tokyo}.
For the purpose of evaluating image retrieval under varying illumination conditions, we define a new evaluation protocol for the \Tokyo dataset. Each image is used in turn as a query. Images from the same location and the same direction (and different illumination conditions) as the query image are deemed positive, while images from different locations are considered negative. Images from the same location as the query image but different direction are excluded from the evaluation, since the overlap between different view directions is not defined. Mean Average Precision (mAP) measure is used to compare the quality of the retrieval with query images excluded from the evaluation, as in~\cite{Radenovic-CVPR18}.

In order to test whether a method still performs well on a `common' day-to-day retrieval task, we evaluate it on the standard revisited Oxford and Paris dataset~\cite{Radenovic-CVPR18}, following the predefined evaluation protocol.

\setlength{\tabcolsep}{3.2pt}
\begin{table}[t] \centering
\begin{tabular}{|l||r||r|r|r|}\hline
    Method & Avg & Tokyo & \roxf & \rpar \\\hline\hline
    Edg+VGG \doubledim [Tab.~\ref{tab:results}] & 71.2 & 85.4 & {\bf 59.4} & 68.8 \\\hline
    Edg+VGG N/D \doubledim & 71.5 & 88.3 & 57.6 & 68.7 \\\hline\hline
    Edg+CLAHE N/D \doubledim & {\bf 72.9} & {\bf 90.5} & 59.1 & {\bf 69.0} \\\hline
    Edg+U-Net jointly N/D \doubledim & 72.3 & 90.0 & 58.1 & 68.8 \\
\hline\multicolumn{5}{c}{}\\[-0.7em]\hline
    Edg+VGG 512 & 70.0 & 81.1 & {\bf 60.1} & 68.9 \\\hline
    Edg+VGG N/D 512 & 71.1 & 85.4 & 59.2 & 68.7 \\\hline\hline
    Edg+CLAHE N/D 512 & {\bf 72.4} & {\bf 88.4} & 59.4 & {\bf 69.3} \\\hline
    Edg+U-Net jointly N/D 512 & 72.1 & 87.8 & 59.8 & 68.7 \\\hline
\end{tabular}\\[3pt]

    \caption{Comparison of ensembles consisting of EdgeMAC ~\cite{Radenovic-ECCV18} (Edg) and VGG GeM~\cite{Radenovic-TPAMI18} (VGG) trained on the Retrieval-SfM-N/D data, without and with the photometric normalisation (first two and second two rows of each block). Whitening is computed from concatenated descriptors and results are reported for the full 1024D (top block) or after dimensionality reduction to 512D (bottom block). For each dimensionality, best score is in bold.}
\label{tab:ensembles}
\end{table}

\subsection{Compared methods}

We compare the performance of the proposed methods with a number of baseline methods. We evaluate on \Tokyo and revisited Oxford and Paris on the Medium protocol. The results are summarised in Tab.~\ref{tab:results}.

The two {\bf baseline methods}, namely VGG GeM baseline and EdgeMAC baseline, are pre-trained networks provided by~\cite{Radenovic-TPAMI18} and~\cite{Radenovic-ECCV18}. For VGG GeM, we copy the scores reported in author's GitHub Page\footnote{https://github.com/filipradenovic/cnnimageretrieval-pytorch} for the PyTorch implementation. For EdgeMAC, we use trained network with Matlab evaluation script from authors' project page\footnote{http://cmp.felk.cvut.cz/cnnimageretrieval/}. In both cases, whitened descriptors were used for comparison. EdgeMAC baseline performs poorly but is shown to enhance image retrieval performance under severe illumination changes~\cite{Radenovic-ECCV18}. Therefore, we further improve the baseline by implementing the idea of~\cite{Chum-CVPR06} to concatenate the descriptors of VGG GeM and EdgeMAC, denoted as EdgeMAC+VGG. The individual descriptors are not whitened separately but instead, a new whitening is computed on the concatenated descriptors. To show the effect of the new dataset without any input data normalisation, we also provide results for VGG GeM fine-tuned on the introduced Retrieval-SfM-N/D dataset.

The impact of {\bf normalisation} is demonstrated through three hand-crafted methods (gamma, histogram equalisation, CLAHE) and three models based on the normalisation network, each trained using a different approach. The first model is trained by fine-tuning the embedding network - a pre-trained normalisation network is used in place of a hand-crafted normalisation with the same training procedure. It can be seen that the pre-trained normalisation network is comparable to the hand-crafted normalisation methods. Next, fine-tuning of the normalisation network is evaluated - VGG GeM is not trained but used solely to provide gradient to the normalisation network. For the last model, both networks were trained jointly.

In Tab.~\ref{tab:ensembles}, {\bf ensemble models} are tested to evaluate the impact of Retrieval-SfM-N/D dataset and photometric normalisation on more complex models. Each ensemble model consists of two networks, VGG GeM and EdgeMAC, which are trained separately. After their descriptors are concatenated, a new whitening is computed on the concatenated descriptors. The final descriptor dimensionality is either full 1024 dimensions or, to enable a fair comparison, is reduced to 512 dimensions. The dimensionality reduction is performed together with the whitening as in~\cite{Radenovic-TPAMI18}, keeping the most discriminative basis for non-matching pairs.

\newcommand{\veryshortarrow}[1][3pt]{\mathrel{%
   \hbox{\rule[\dimexpr\fontdimen22\textfont2-.2pt\relax]{#1}{.4pt}}%
   \mkern-4mu\hbox{\usefont{U}{lasy}{m}{n}\symbol{41}}}}
\setlength{\tabcolsep}{3.1pt}
\begin{table}[t] \centering
\begin{tabular}{|l||r|r|r|r|r|r|} \hline
    \multirow{2}{*}{Method} & \multicolumn{2}{c|}{\footnotesize \{day, sunset\}} & \multicolumn{2}{c|}{\footnotesize \{sunset, night\}} & \multicolumn{2}{c|}{\footnotesize \{day, night\}}\\
    &D$\veryshortarrow$S & S$\veryshortarrow$D & \hspace{4pt}S$\veryshortarrow$N & N$\veryshortarrow$S & D$\veryshortarrow$N & N$\veryshortarrow$D \\\hline\hline

    VGG GeM~\cite{Radenovic-TPAMI18} & 95.7 & 97.5 & 71.2 & 73.0 & 62.0 & 67.3 \\\hline
    VGG GeM N/D & 96.5 & 97.1 & 74.7 & 80.3 & 67.6 & 74.8 \\\hline
    EdgeMAC+VGG \doubledim & 97.2 & 97.7 & 79.5 & 80.6 & 73.5 & 74.9 \\\hline\hline

    CLAHE N/D & 96.5 & 97.5 & 79.7 & 86.9 & 72.5 & 81.3 \\\hline
    U-Net embed N/D & 96.6 & 97.1 & 78.5 & 86.1 & 70.9 & 80.4 \\\hline
    U-Net norm N/D & 97.0 & 97.5 & 75.2 & 79.5 & 66.9 & 72.9 \\\hline
    U-Net jointly N/D & 96.8 & 97.8 & 79.6 & 84.8 & 71.6 & 79.8 \\\hline

\end{tabular}\\[3pt]

    \caption{The performance (measured by mAP) for chosen methods from Tab.~\ref{tab:results} on the \Tokyo dataset for different lighting conditions of the query and retrieved pair of images. Each column corresponds to the query image being taken either during day (D), sunset (S) or night (N) and the retrieved image being taken during one of the remaining two.}
\label{tab:titech_classes}
\end{table}

\subsection{Retrieval results}
The retrieval results are summarised in Tab.~\ref{tab:results}. Methods followed by ``ND'' were trained using a mixture of Retrieval-SfM and Retrieval-SfM-N/D datasets with the ratio 3:1 respectively, while other methods were trained using Retrieval-SfM only. Methods with a citation were taken from publicly available sources.

\paragraph{(i)} All image normalisation methods outperform the baseline methods with the same descriptor dimensionality (VGG GeM and EdgeMAC) on the \Tokyo dataset by a large margin, see Tab.~\ref{tab:results}. 
Combining VGG GeM and EdgeMAC descriptors delivers satisfactory results on the \Tokyo dataset at the cost of an increased descriptor dimensionality. However, the performance of the concatenated descriptors is slightly decreased on the Oxford and Paris datasets. 

\paragraph{(ii)} The effect of the newly introduced dataset, Retrieval-SfM-N/D, is visible in both cases, without the normalisation step -- comparing \mdl{VGG GeM} and \mdl{VGG GeM N/D}, and with the normalisation step -- comparing \mdl{CLAHE} and \mdl{CLAHE N/D}, or \mdl{U-Net jointly} and \mdl{U-Net jointly N/D} methods.

\paragraph{(iii)} An embedding network with no photometric normalisation fine-tuned on the novel dataset \mdl{VGG GeM N/D} performs better than the baseline \mdl{VGG GeM}, but is still inferior to methods using a photometric normalisation.

\paragraph{(iv)} The two best performing methods -- \mdl{CLAHE N/D} and \mdl{U-Net jointly N/D} -- preform similarly on all datasets, and are closely followed by another three methods \mdl{U-Net norm N/D}, \mdl{Histogram eq. N/D} and \mdl{CLAHE}.

\paragraph{(v)} Performance can be further increased by creating an ensemble of VGG GeM and EdgeMAC.
In all cases, methods trained with the proposed Retrieval-SfM-N/D dataset outperform comparable methods that do not use it. Similarly, the photometric normalisation always improves the results even when combined with EdgeMAC.

\vspace{.5\baselineskip} From the experiments, we conclude that the photometric normalisation significantly improves the performance (i), and that training the network on image pairs exhibiting illumination changes, such as Retrieval-SfM-N/D, is important (ii). The photometric normalisation enhances visual information that is difficult to capture for the embedding network alone, even when trained on data exhibiting illumination changes (iii). The currently proposed learnable photometric normalisation does not provide additional information over the CLAHE normalisation, that cannot be extracted later by the embedding network (iv). This is supported by the fact that freezing the normalisation network pre-trained for a different task (\mdl{U-Net embed N/D}) is beneficial for retrieval result on \Tokyo, comparably to \mdl{U-Net jointly N/D}.

We further analyse the performance on the \Tokyo dataset with respect to different light conditions of the query and retrieved images by breaking down the dataset illumination types. In Tab. \ref{tab:titech_classes}, we provide results for the six available combinations query-type $\to$ database-type, such as a night query retrieving a day image (denoted as N$\veryshortarrow$D).

It can be seen that the lowest scores are obtained for day-night image pairs, followed by sunset-night image pairs where the query is either one of the pair. For those four cases, presented methods bring the largest improvement.

\section{Conclusions}

In this work, we proposed a photometric normalisation step for image retrieval under varying illumination conditions. We have experimentally shown that such a normalisation significantly improves the performance in the presence of significant illumination changes, while preserving the state-of-the-art performance in similar illumination conditions. We have compared several methods, both hand-crafted and learnable. The best performing methods based on CLAHE and on the proposed learned normalisation with the U-Net architecture perform similarly well, while the hand-crafted method being significantly faster. 
Further, we have constructed a novel dataset Retrieval-SfM-N/D. The importance of fine-tuning the network on training data that exhibit significant changes in illumination was shown.

\paragraph{Acknowledgments.}
This work was supported by the GA\v{C}R grant 19-23165S and the CTU student grant SGS17/185/OHK3/3T/13.

\end{document}